\documentclass[conference]{IEEEtran}

\usepackage{cite}
\usepackage{amsmath,amssymb,amsfonts}
\usepackage{algorithmic}
\usepackage{graphicx}
\usepackage{textcomp}
\usepackage{xcolor}
\usepackage{comment}
\usepackage{booktabs}
\usepackage{tikz}
\usepackage{url}

\def\BibTeX{{\rm B\kern-.05em{\sc i\kern-.025em b}\kern-.08em T\kern-.1667em\lower.7ex\hbox{E}\kern-.125emX}}

\usetikzlibrary{arrows.meta, positioning, fit, calc, backgrounds,
                 decorations.pathreplacing}
\usepackage[T1]{fontenc}
\IEEEoverridecommandlockouts
\begin{document}

\title{Learning Ad Hoc Network Dynamics via Graph-Structured World Models\thanks{This work is supported by The Scientific and Technological Research Council of Türkiye (TUBITAK) 1515 Frontier R{\&}D Laboratories Support Program for Türk Telekom 6G R{\&}D Lab under project number 5249902.}}
\author{
\IEEEauthorblockN{
Can Karacelebi\IEEEauthorrefmark{1},
Yusuf Talha Sahin\IEEEauthorrefmark{1}\IEEEauthorrefmark{2},
Elif Surer\IEEEauthorrefmark{3},
Ertan Onur\IEEEauthorrefmark{1}
}
\IEEEauthorblockA{\IEEEauthorrefmark{1}Department of Computer Engineering, Middle East Technical University, Ankara, Turkiye}
\IEEEauthorblockA{\IEEEauthorrefmark{2}Türk Telekom, Ankara, Turkiye}
\IEEEauthorblockA{\IEEEauthorrefmark{3}Graduate School of Informatics, Middle East Technical University, Ankara, Turkiye}
\IEEEauthorblockA{
\IEEEauthorrefmark{1}\{can.karacelebi, eronur\}@metu.edu.tr \quad
\IEEEauthorrefmark{2}yusuftalha.sahin@turktelekom.com.tr \quad
\IEEEauthorrefmark{3}elifs@metu.edu.tr
}
}

\maketitle

\begin{abstract}
Ad hoc wireless networks exhibit complex, innate and coupled dynamics: node mobility, energy depletion and topology change that are difficult to model analytically. Model-free deep reinforcement learning requires sustained online interaction whereas existing model based approaches use flat state representations that lose per node structure. Therefore we propose G-RSSM, a graph structured recurrent state space model that maintains per node latent states with cross node multi head attention to learn the dynamics jointly from offline trajectories. We apply the proposed method to the downstream task clustering where a cluster head selection policy trains entirely through imagined rollouts in the learned world model. Across 27 evaluation scenarios spanning MANET, VANET, FANET, WSN and tactical networks with N=30 to 1000 nodes, the learned policy maintains high connectivity with only trained for N=50. Herein, we propose the first multi physics graph structured world model  applied to combinatorial per node decision making in size agnostic wireless ad hoc networks.
\end{abstract}


\section{Introduction}
Since the concept of ad hoc wireless networks are coined, resources and efficiency have been the main aspects of managing functioning networks. Core dynamics of the Internet had contrasting characteristics when ad hoc networks were deployed and decentralized structures were heavily used for better management in order to supply efficient routing or interference mitigation. This grounds the core characteristics of ad hoc wireless networks: being governed by coupled, nonlinear dynamics within itself. Node mobility affects topology while energy consumption leads node death leading to suspended connectivity or decreased communication quality. Ad hoc network optimization has been a spotlight research including clustering, routing and scheduling requiring accurate predictions of the given network's evolution under different control decisions. Analytical and formal methods capture isolated aspects (path loss, mobility) but not the coupled system itself. Applications on simulation environments become expensive and non-differentiable. This brings the question: Is it possible to learn a compact, differentiable model of the entire network dynamics?

World models \cite{ha2018WM} learn environment dynamics in latent space (usually obtained with an encoder through observations) and train policies through imagined rollouts. These approaches proved to be competitive and achieved state-of-the art performance across more than 150 diverse domains \cite{hafner2024dreamer}. A key benefit is that, once offline trajectories are collected, the policy trains entirely on imagined rollouts within the world model, no further interaction with the real network is required. There are recent works which have begun with applications of world models to wireless networks domain: flat RSSM (Recurrent State Space Model) for V2X scheduling \cite{wang2025worldmodelbasedlearninglongterm}, Wireless Dreamer for wireless edge intelligence optimization \cite{zhao2025worldmodelscognitiveagents} and dual-mind world models for access scheduling \cite{wang2025dualWM}. Meanwhile, graph-structured recurrent models with latent variables have been explored for link prediction \cite{vgrnn2019}, multi object forecasting and concurrently for building HVAC control \cite{berkes2026graph}. Nevertheless, these methods have not been applied to wireless network dynamics with per-node action conditioning.
The following open problems remain within the pioneering world model approaches for ad hoc networks: all existing world models use flat Dreamer like RSSM losing per-node structure, none addresses combinatorial decentralized per-node decisions ($\{0,1\}^N$) and the focus on generalization across varying network sizes.

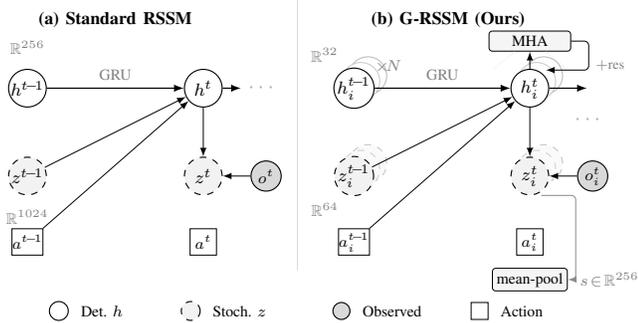
\begin{figure}[t]
      \centering
      \resizebox{\columnwidth}{!}{%
          \begin{tikzpicture}[
    >=latex,
    det/.style={circle, draw=black, fill=white, line width=0.5pt,
                minimum size=0.6cm, font=\footnotesize, inner sep=0pt},
    stoch/.style={circle, draw=black, dashed, fill=black!5, line width=0.5pt,
                  minimum size=0.6cm, font=\footnotesize, inner sep=0pt},
    observed/.style={circle, draw=black, fill=black!15, line width=0.5pt,
                     minimum size=0.48cm, font=\scriptsize, inner sep=0pt},
    actnode/.style={rectangle, draw=black, fill=white, line width=0.4pt,
                    minimum size=0.42cm, font=\scriptsize, inner sep=0pt},
    modbox/.style={rectangle, draw=black, fill=black!4, line width=0.4pt,
                   rounded corners=1.5pt, minimum height=0.32cm,
                   font=\scriptsize, inner sep=2pt},
    arr/.style={->, line width=0.4pt},
    thickarr/.style={->, line width=0.55pt},
    dlbl/.style={font=\scriptsize, text=black!50},
]

\node[font=\footnotesize\bfseries] at (1.4, 3.3) {(a) Standard RSSM};

\node[det] (h0) at (0, 2.2) {$h^{t\!-\!1}$};
\node[stoch] (z0) at (0, 0.8) {$z^{t\!-\!1}$};
\node[actnode] (a0) at (0, -0.25) {$a^{t\!-\!1}$};

\node[det] (h1) at (2.8, 2.2) {$h^t$};
\node[stoch] (z1) at (2.8, 0.8) {$z^t$};
\node[observed] (o1) at (3.8, 0.8) {$o^t$};
\node[actnode] (a1) at (2.8, -0.25) {$a^t$};

\draw[thickarr] (h0) -- node[above, dlbl] {GRU} (h1);
\draw[arr] (z0) -- (h1);
\draw[arr] (a0) -- (h1);
\draw[arr] (h1) -- (z1);
\draw[arr] (o1) -- (z1);

\node[dlbl, above=3pt of h0] {$\mathbb{R}^{256}$};
\node[dlbl, below=3pt of z0] {$\mathbb{R}^{1024}$};

\draw[thickarr] (h1) -- ++(0.6, 0);
\node[font=\normalsize, text=black!30] at (3.75, 2.2) {$\cdots$};

\draw[black!15, line width=0.3pt] (4.3, -0.7) -- (4.3, 3.6);

\begin{scope}[xshift=4.7cm]
\node[font=\footnotesize\bfseries] at (2.0, 3.3) {(b) G-RSSM (Ours)};

\node[det] (hi0) at (0.5, 2.2) {$h_i^{t\!-\!1}$};
\node[stoch] (zi0) at (0.5, 0.8) {$z_i^{t\!-\!1}$};
\node[actnode] (ai0) at (0.5, -0.25) {$a_i^{t\!-\!1}$};

\begin{scope}[on background layer]
    \node[det, draw=black!25, fill=white] at (0.72, 2.4) {};
    \node[det, draw=black!20, fill=white] at (0.61, 2.31) {};
    \node[stoch, draw=black!22, fill=black!2] at (0.72, 1.0) {};
    \node[stoch, draw=black!18, fill=black!2] at (0.61, 0.91) {};
\end{scope}
\node[dlbl] at (1.05, 2.55) {$\times\!N$};

\node[det] (hi1) at (3.3, 2.2) {$h_i^t$};
\node[stoch] (zi1) at (3.3, 0.8) {$z_i^t$};
\node[observed] (oi1) at (4.3, 0.8) {$o_i^t$};
\node[actnode] (ai1) at (3.3, -0.25) {$a_i^t$};

\begin{scope}[on background layer]
    \node[det, draw=black!25, fill=white] at (3.52, 2.4) {};
    \node[det, draw=black!20, fill=white] at (3.41, 2.31) {};
    \node[stoch, draw=black!22, fill=black!2] at (3.52, 1.0) {};
    \node[stoch, draw=black!18, fill=black!2] at (3.41, 0.91) {};
\end{scope}

\draw[thickarr] (hi0) -- node[above, dlbl] {GRU} (hi1);
\draw[arr] (zi0) -- (hi1);
\draw[arr] (ai0) -- (hi1);
\draw[arr] (hi1) -- (zi1);
\draw[arr] (oi1) -- (zi1);

\node[dlbl, above left=2pt and -3pt of hi0] {$\mathbb{R}^{32}$};
\node[dlbl, below left=2pt and -3pt of zi0] {$\mathbb{R}^{64}$};

\node[modbox, minimum width=1.3cm] (mha) at (3.3, 2.95) {MHA};
\draw[arr] (hi1.north) -- (mha.south);

\draw[arr, rounded corners=2pt]
    (mha.east) -- ++(0.28, 0) |- ([xshift=1pt,yshift=2pt]hi1.north east);

\node[dlbl, scale=0.9] at (4.58, 2.56) {$+$res};

\draw[densely dotted, line width=0.2pt, black!30]
    ([xshift=-0.25cm]mha.south) -- ++(-0.35, -0.3);
\draw[densely dotted, line width=0.2pt, black!30]
    ([xshift=0.25cm]mha.south) -- ++(0.35, -0.3);

\node[modbox] (pool) at (3.3, -0.85) {mean-pool};
\draw[arr, black!45, rounded corners=2pt]
    ([xshift=0.18cm]zi1.south) -- ++(0.55,0) |- (pool.east);
\node[dlbl, right=2pt of pool] {$s\!\in\!\mathbb{R}^{256}$};

\draw[thickarr] (hi1.east) -- ++(0.6, 0);
\node[font=\normalsize, text=black!30] at (4.25, 1.7) {$\cdots$};
\end{scope}

\node[det, minimum size=0.3cm] (l1) at (0.5, -1.35) {};
\node[right=2pt of l1, font=\scriptsize] {Det.\ $h$};

\node[stoch, minimum size=0.3cm] (l2) at (2.6, -1.35) {};
\node[right=2pt of l2, font=\scriptsize] {Stoch.\ $z$};

\node[observed, minimum size=0.27cm] (l3) at (5.0, -1.35) {};
\node[right=2pt of l3, font=\scriptsize] {Observed};

\node[actnode, minimum size=0.27cm] (l4) at (7.2, -1.35) {};
\node[right=2pt of l4, font=\scriptsize] {Action};

\end{tikzpicture}
      }
      \caption{(a) Standard RSSM compresses $N$ nodes into a single state vector.
      (b) G-RSSM maintains per-node states with cross node attention.}
      \label{fig:rssm_comparison}
  \end{figure}

We propose a graph structured Recurrent State Space Model: G-RSSM that maintains per-node recurrent latent states with cross node multi head attention. Unlike the flat RSSM (Fig.~\ref{fig:rssm_comparison}) which compresses $N$ nodes into a single vector, G-RSSM preserves individiual node dynamics while modeling inter-node interactions.

The contributions of this paper are three-fold: (i) Multi physics graph structured world model for ad hoc networks. We introduce a per-node RSSM with cross node attention that jointly learns five coupled network processes: node mobility, energy consumption, topology evaluation, clustering reward and network degradation (a learned continue predictor) in a unified graph structured latent space, (ii) Imagination-based combinatorial optimization for wireless networks. We are the first to apply world model rollouts to per-node binary decision-making on wireless networks. The learned continue predictor estimates the probability that the network remains operational at each imagined step. When the predictor signals impending degradation (e.g. energy depletion causing node death), the rollout truncates, discounting future returns and teaching the policy to avoid actions that accelerate network collapse. Each node's previous action decision feeds back into its dynamics step enabling the world model to capture how individual clustering decisions propagate through the network, and (iii) Size agnostic learned network dynamics. The learned dynamics model itself generalizes to unseen network sizes. When trained on smaller sizes, it can model dynamics for larger networks without retraining. This emerges from shared weight per node GRU (Gated Recurrent Unit) + attention. The code used in this work\footnote{\url{https://github.com/cankaracelebi/WM-cluster}} is publicly available.

\section{Background and Problem Formulation}

While model-free reinforcement learning (RL) is mostly employed in the domain of applied artificial intelligence (AI) to the network operations, model-based RL evolved to prove its competitive performance and unique benefits. Model based RL aims to learn the internal dynamics of the presumed underlying Markov Decision Process (MDP). Model-based RL learns dynamics $\hat{p}(s_{t+1} | s_t, a_t)$ from collected experience and aims to train a policy via imagined trajectories directly from the learned dynamics \cite{moerland2022modelbasedreinforcementlearningsurvey}. The key advantage over model-free methods is being substantially more sample efficient. Policy training is conducted without real interaction to the proposed environment.

\subsection{World Models and Recurrent State Space Models}
The idea of model-based RL is extended to increase the accuracy of the model itself with the emergence of World models \cite{ha2018WM}. Original world model idea proposed a VAE (Variational Auto Encoder) with a recurrent dynamics module and a compact controller in order to train entirely in the famously called "dream". PlaNet\cite{hafner2019learninglatentdynamicsplanning} introduced Recurrent State Space Model (RSSM) to capture long-term dependencies with a stochastic latent to capture the stochastic behavior of the environment. RSSM structure can be decomposed into a four arm structure for interpretability:
$$ \textbf{Dynamics: } h_t = f_{\phi}(h_{t-1}, z_{t-1}, a_{t-1}) \textit{ (GRU)}, $$
$$ \textbf{Prior: } p_{\phi}(z_t|h_t)  \textit{ (Predict latent from dynamics)},$$
$$ \textbf{Posterior: }  q_{\phi}(z_t|h_t, o_t) \textit{ (update with observation)},$$
$$ \textbf{Decode: } p_{\phi}(o_t, r_t | h_t, z_t) \textit{ (Reconstruction)}$$
where $h_t$ is the deterministic current state and $z_t$ is the stochastic latent space. Therefore, in a given time, the full latent representation is $s_t = (h_t, z_t)$ as a surrogate to the state $s_t$. $h_t$ is inferred from a recurrent information capable model (RNN, GRU) and $z_t$ represents the uncertainty as a stochastic estimator. Using a recurrent powered neural network $f_{\phi}$ computes the deterministic transition for the new hidden state (memory) $h_t$ using given past memory $h_{t-1}$, previous stochastic state $z_{t-1}$ and the previous action $a_{t-1}$. Given the recurrent memory $h_t$ prior $p_{\phi}$ predicts the next latent space $z_t$. After experiencing a real observation $o_t$, a better latent representation is inferred via the posterior $q_{\phi}(z_t | h_t, o_t)$ as a belief update. Posterior helps with correct inference of the next latent space given the current observation, prior enables to guess before looking at the current observation making it possible to imagine the future with the dynamics model learned itself. With an accurate prior model, imagined rollouts will successfully represent the real dynamics enabling planning in latent space in a future horizon. This architecture is widely used as a base to form world model based RL methods. Dreamer \cite{hafner2020dreamcontrollearningbehaviors} proposed an actor-critic trained entirely from RSSM rollouts via analytic gradients. DreamerV3 \cite{hafner2024dreamer} as a sucessor to Dreamer model family proposed categorical latents, symlog predictions and a continue predictor to master diverse tasks with a single configuration. RSSM compresses entire environment into single $(h_t, z_t)$, for a network of N nodes, per node information is also compressed preventing the model to distinguish which node is depleting energy or responsible for losing connectivity.

\subsection{Graph Structured State Space Models}

Graph structured neural networks and state space models with latent variables have been used in varying tasks to preserve the innate graph information. R-SSM \cite{yang2020relationalstatespacemodelstochastic} proposes a GNN (Graph Neural Network) + hierarchical SSM with normalizing flows for multi object forecasting. VGRNN \cite{hajiramezanali2020variationalgraphrecurrentneural} proposed a per-node recurrent model with a per-node latent for dynamic link prediction, used for graph generation not control, no imagination nor policy training is designed. Concurrently, Graph Dreamer architecture is utilized to build HVAC control \cite{berkes2026graph}. We adopt this architectural paradigm for wireless network dynamics where the key challenges are per node action conditioning (individual decision affects its dynamics), multi-physics joint prediction (mobility, energy, topology and degradation) and combinatorial action spaces ($\{0,1\}^N$).

\subsection{GNNs for Wireless Networks}

GNNs allow building natural model relational structure on ad hoc networks. Network nodes are modeled as graph nodes and edges as communication links. A recent architecture facilitating attention is GATv2\cite{brody2022attentivegraphattentionnetworks} which proposes dynamic attention,  a modification over GAT \cite{velickovic2018graphattentionnetworks} mitigating expressivity issues and limitations. GNN based methods are widely used as policy networks for power control, scheduling and resource management in different domains. VAST-GCN utilized GNNs for a supervised clustering framework for vehicular ad hoc networks. Although GNNs are used as function approximators, similarly to multi-layer perceptrons (MLPs) in applied RL methods,they are not fully utilized as dynamic models (world models. Within the wireless ad hoc networks there is no prior combination of GNN message passing with RSSM temporal dynamics. Another key advantage of GNN architectures is that they are size agnostic by design, same GNN weights work on any N with competitive performance \cite{naderializadeh2020wirelesspowercontrolcounterfactual}.

\subsection{Network Model}
Network model is designed within alignment of general physics and internal dynamics with robust characteristics. We define a network with $N$ mobile nodes in $[0,L]^2$, a node $i$ at time t has position $p_i^{t}$, velocity $v_i^{t}$, energy $e_i^{t}$ and a binary status (CH status in our experiments) $c_i^{t} \in \{0,1\}$. We model the energy as $E_{tx}(d) = E_{elec}$ + $\varepsilon _{amp} \cdot d^2$ ($E_{elec}$: base electric cost to transmit, $\varepsilon_{amp}$: amplifier cost) with additional CH overhead proportional to the cluster size ($\propto |C_i|$) and an idle drain $E_{idle}/step$ \cite{heinzelman}. Death of a node due to energy is defined as $e_i^t \leq E_{death}$ in a given $t$ for node $i$. We model the channel with respect to simple path loss model as $P_{r,ij} = k_o P_t(\frac{d_0}{d_{ij}})^\eta $ where $k_0$ is the reference path loss factor, $\eta$ is the path loss exponent ($\eta$ = 3) and $d_0$ is the reference distance. A network is formally defined as a graph $G_t = (V, E_t)$ where edge $(i,j) \in E_t$ exist if and only if $P_{r,ij} \geq \gamma_{rx}$. This graph changes every timestep due to mobility and node energy availability.

\subsection{Clustering as a Proof-of-Concept Application}

Clustering is a conceptual method to appoint semi-centralized CH nodes to manage routing for the nodes in its vicinity in order to decrease the cumulative control overhead. Clustering is favored for scenarios to optimize energy consumption for network lifetime longevity. We define a centralized method for each node for their roles as combinatorial CH selection $a_t \in \{0,1\}^N$ designed to be solved each timestep. Reward is formulated as a physics-based multi-objective reward $r_t = w_sR_{stability} + w_{e}R_{energy} +w_{c}R_{connectivity}+ w_{h}R_{CH} + w_{\tau}R_{temp} - w_{p} \cdot1[u > \theta]$ where u is the unclustered fraction of nodes and $\theta$ is a chosen coverage threshold. $R_{CH}$ is a penalty term for too many or few CHs (target $\sim \sqrt{N}$) and $R_{temp}$ is designed as the temporal term penalty for frequent CH changes. Overall formulation forms a suitable testbed for world models because of coupled multi-objective, sequential decisions with long horizon effects (e.g., energy depletion) and the importance of the graph structure. We define the network lifetime as the number of timesteps until the connectivity falls under $50 \%$.

\section{G-RSSM: A Graph Structured World Model}
\begin{figure*}[t] \centering \resizebox{0.77\textwidth}{!}{%
\begin{tikzpicture}[
    >=Stealth,
    blk/.style={rectangle, draw=black, line width=0.5pt, rounded corners=1.5pt,
        minimum height=0.48cm, minimum width=2.0cm, font=\scriptsize,
        align=center, inner sep=2pt, fill=white},
    sm/.style={rectangle, draw=black, line width=0.4pt, rounded corners=1pt,
        minimum height=0.38cm, minimum width=0.9cm, font=\tiny,
        align=center, inner sep=1.5pt, fill=white},
    io/.style={blk, fill=black!5, line width=0.55pt},
    a/.style={-{Stealth[length=2pt,width=1.6pt]}, line width=0.45pt},
    ab/.style={-{Stealth[length=2.5pt,width=2pt]}, line width=0.55pt},
    ad/.style={-{Stealth[length=2pt,width=1.6pt]}, dashed, line width=0.4pt},
    aloop/.style={-{Stealth[length=2pt,width=1.6pt]}, densely dash dot, line width=0.45pt},
    d/.style={font=\tiny, text=black!50},
]

\node[io, minimum width=1.8cm] (inp) at (0, -1.2)
    {$\mathcal{G}^t$};
\node[d, below=-0.01cm of inp] {$N{\times}7$};

\node[blk, below=0.4cm of inp, minimum width=1.8cm] (enc)
    {GATv2 Encoder};
\node[d, below=-0.01cm of enc] {2-layer, 4-head};
\draw[ab] (inp) -- (enc);

\node[blk, below=0.4cm of enc, minimum width=1.8cm] (oemb)
    {$\mathbf{o}_i \!\in\! \mathbb{R}^{64}$};
\node[d, below=-0.01cm of oemb] {$\times\!N$ nodes};
\draw[a] (enc) -- (oemb);

\begin{scope}[on background layer]
    \node[fill=black!2, rounded corners=3pt,
          fit=(inp)(enc)(oemb), inner sep=6pt] (ez) {};
\end{scope}
\node[font=\tiny\bfseries, anchor=south]
    at ($(ez.north)+(0,0.02)$) {Encoder};

\node[blk, minimum width=2.4cm] (gru) at (4.8, 0)
    {Per-node GRU(32)};

\node[blk, below=0.4cm of gru, minimum width=2.4cm] (mha)
    {Cross-node MHA};
\node[d, below=0.06cm of mha] {4-head self-attn + LN};
\draw[a] (gru) -- (mha);

\node[sm, below left=0.45cm and 0.15cm of mha] (pri) {Prior $p$};
\node[sm, below right=0.45cm and 0.15cm of mha] (pos) {Post.\ $q$};
\draw[a] (mha.south) -- ++(0,-0.12) -| (pri.north);
\draw[a] (mha.south) -- ++(0,-0.12) -| (pos.north);

\node[font=\tiny] (kl) at ($(pri.south)!0.5!(pos.south)+(0,0.15)$)
    {$D_{\mathrm{KL}}$};

\node[blk, below=0.06cm of kl, minimum width=2.0cm] (zcat)
    {$\mathbf{z}_i \!\sim\! \mathrm{Cat}(8{\times}8)$};
\draw[a] (pri.south) -- ++(0,-0.03) -| ([xshift=-0.2cm]zcat.north);
\draw[a] (pos.south) -- ++(0,-0.03) -| ([xshift=0.2cm]zcat.north);

\begin{scope}[on background layer]
    \node[fill=black!3, rounded corners=3pt,
          fit=(gru)(mha)(pri)(pos)(zcat)(kl), inner sep=6pt] (rz) {};
\end{scope}
\node[font=\tiny\bfseries, anchor=south west]
    at ($(rz.north west)+(0,0.02)$) {G-RSSM (${\times}N$)};

\draw[ab] (oemb.east) -- ++(0.4,0) |- ([yshift=0.1cm]gru.west)
    node[pos=0.75, above, d] {$N{\times}64$};

\coordinate (rec_x) at ($(rz.west)+(-0.4,0)$);
\draw[aloop] (zcat.west) -- (zcat.west -| rec_x)
    -- (gru.west -| rec_x) -- (gru.west);
\node[d, rotate=90, anchor=south]
    at ($(rec_x)+(0.0,-0.1)$) {$\mathbf{z}_i^{t\text{-}1},a_i^{t\text{-}1}$};

\node[blk, below=0.5cm of zcat, minimum width=2.4cm] (agg)
    {Mean-pool $\to$ $\mathbf{s}^t\!\in\!\mathbb{R}^{256}$};
\draw[ab] (zcat) -- (agg) node[midway, right, d] {$N{\times}96$};

\node[blk, below=0.35cm of agg, minimum width=2.6cm, font=\tiny] (dec)
    {Pos $|$ En $|$ Adj $|$ $\hat{r}_t$ $|$ $\hat{c}_t$};
\node[d, below=-0.01cm of dec] {\itshape train only};
\draw[a] (agg) -- (dec);

\node[font=\scriptsize, below=0.2cm of dec] (loss)
    {$\mathcal{L}_{\mathrm{WM}} \!=\!
      \mathcal{L}_{\mathrm{recon}} + \beta\max(D_{\mathrm{KL}},\eta)$};

\node[blk, fill=black!4, minimum width=2.2cm,
      right=1.6cm of mha] (nenc)
    {Node encoder};
\node[d, below=-0.08cm of nenc]
    {$[\mathbf{h}_i,\mathbf{z}_i,a_i^{t\text{-}1}]\!\to\!64$};

\node[blk, fill=black!4, minimum width=2.2cm,
      below=0.35cm of nenc] (aatt)
    {Self-attention};
\node[d, below=-0.01cm of aatt] {4-head MHA + LN};
\draw[a] (nenc) -- (aatt);

\node[blk, fill=black!4, minimum width=2.2cm,
      below=0.35cm of aatt] (amlp)
    {Per-node MLP};
\node[d, below=-0.01cm of amlp] {$64\!\to\!128\!\to\!1$};
\draw[a] (aatt) -- (amlp);

\node[blk, below=0.35cm of amlp, minimum width=2.2cm] (bern)
    {$\pi(a_i{=}1) = \sigma(\ell_i)$};
\draw[a] (amlp) -- (bern);

\node[io, below=0.35cm of bern, minimum width=2.2cm] (chout)
    {$\mathbf{a}^t \!\in\! \{0,1\}^N$};
\draw[ab] (bern) -- (chout);

\node[blk, below=0.4cm of chout, minimum width=2.2cm] (critic)
    {Critic $V(\mathbf{s}^t)$};
\node[d, below=-0.01cm of critic] {MLP $\to \mathbb{R}$};

\begin{scope}[on background layer]
    \node[fill=black!4, rounded corners=3pt,
          fit=(nenc)(aatt)(amlp)(bern)(chout)(critic), inner sep=6pt] (az) {};
\end{scope}
\node[font=\tiny\bfseries, anchor=south]
    at ($(az.north)+(0,0.02)$) {Actor-Critic};

\draw[ab] (mha.east) -- (nenc.west)
    node[midway, above, d] {$[\mathbf{h}_i,\mathbf{z}_i]$}
    node[midway, below, d] {$N{\times}96$};

\draw[ad] (chout.east) -- ++(0.35,0) |- (nenc.east)
    node[pos=0.18, right, d] {$a_i^{t\text{-}1}$};

\coordinate (img_y) at ($(loss.south)+(0,-0.4)$);
\coordinate (img_x) at ($(rec_x)+(-0.45,0)$);
\coordinate (gru_entry) at ([yshift=-0.12cm]gru.west);

\draw[aloop, line width=0.55pt]
    (chout.west) -- (az.west |- chout)               
    -- (az.west |- img_y)                             
    -- (img_x |- img_y)                               
    -- (img_x |- gru_entry)                           
    -- (gru_entry);                                    

\node[font=\tiny\bfseries, anchor=north]
    at ($(img_x |- img_y)!0.5!(az.west |- img_y)+(0,0.04)$)
    {imagination: $\mathbf{a}^t \to$ dynamics};

\end{tikzpicture} } \caption{Overall Architecture.} \label{fig:arch} \end{figure*}
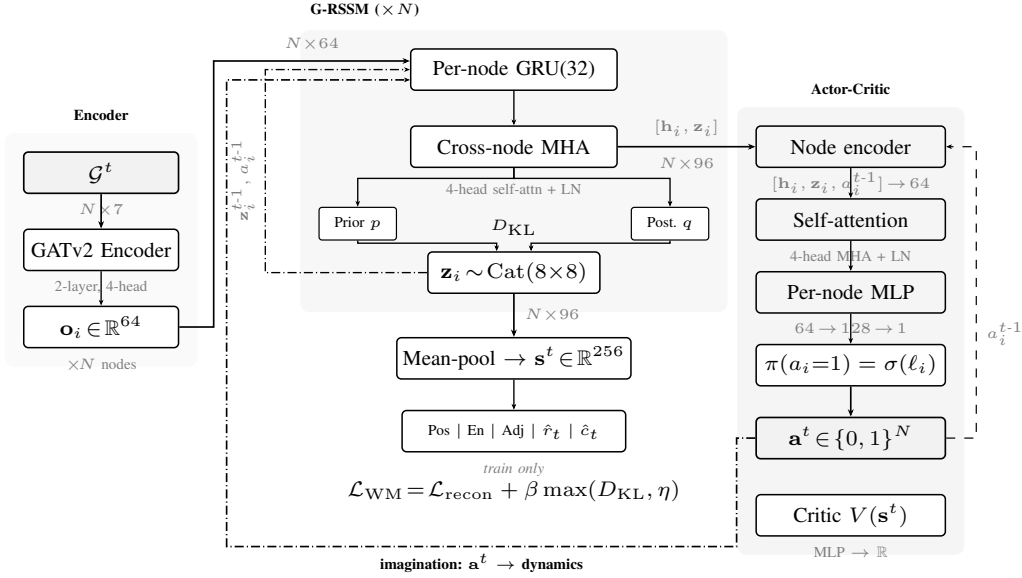

  \begin{table}[b]                                                                                                                                 
    \centering                                                                     
    \caption{Decoder heads from global state $\mathbf{s} \in \mathbb{R}^{256}$.}                                                                                                                
    \label{tab:decoders}                                                                                                                                                                        
    \small                                                                                                                                                                                      
    \resizebox{\columnwidth}{!}{%
    \begin{tabular}{@{}lll@{}}                                                                                                                                                                  
    \toprule
    \textbf{Head} & \textbf{Computation} & \textbf{Loss} \\                                                                                                                                     
    \midrule                                                                                                                                                                                    
    Shared & $\{\mathbf{e}_i\}_{i=1}^{N} = \mathrm{reshape}(\mathrm{MLP}(\mathbf{s})) $ & -- \\
    \midrule                                                                                                                                                                                    
    Position & $\hat{\mathbf{p}}_i = \mathrm{MLP}(\mathbf{e}_i) \in \mathbb{R}^2$ & $\|\hat{\mathbf{p}} - \mathrm{symlog}(\mathbf{p})\|^2$ \\
    Energy   & $\hat{e}_i = \mathrm{MLP}(\mathbf{e}_i) \in \mathbb{R}$ & $\|\hat{e} - \mathrm{symlog}(e)\|^2$ \\                                                                                
    Adjacency & $\hat{A}_{ij} = (\mathbf{W}\mathbf{e}_i)^\top (\mathbf{W}\mathbf{e}_j)$ & $\mathrm{BCE}(\hat{A}, A)$ \\                                                                         
    \midrule                                                                                                       
    Reward   & $\hat{r} = \mathrm{MLP}(\mathbf{s}) \in \mathbb{R}$ & $\|\hat{r} - \mathrm{symlog}(r)\|^2$ \\                                                                                    
    Continue & $\hat{\gamma} = \sigma(\mathrm{MLP}(\mathbf{s})) \in [0,1]$ & $\mathrm{BCE}(\hat{\gamma}, \gamma)$ \\                                                                            
    \bottomrule                                                                                                                                                                                 
    \end{tabular}%
    }                                                                                                                                                                                   
  \end{table}     
We propose G-RSSM as a general architecture for learning graph-structured dynamics. We tailored G-RSSM to be intuitively aligned with wireless ad hoc networks empowering decentralized planning and individual actions. The overall architecture is shown in Fig.~\ref{fig:arch}.

\subsection{Architecture Overview}
We extend RSSM from a single global state to $N$ per node states with cross node communication. The GATv2 encoder processes the network graph into per node observations. The G-RSSM dynamics maintain per node deterministic states (GRU) with cross node attention and per node stochastic states (categorical). Decoders reconstruct the network state for training but during inference only the encoder and dynamics are needed.

\subsection{GNN Encoder}
We form a GNN based GATv2 encoder
\begin{equation}
    \mathbf{H}^t = \mathrm{GATv2}_{\mathrm{2-layer}}(\mathbf{X^t, \mathbf{E^t}}) \in \mathbb{R}^{N \times 64}
\end{equation}
with a 7 dimensional input per node: position, velocity, energy, degree, CH status, alive and a unique cluster number (id). The edge features consist of the simple path loss model. We form a two-layer GATv2 with four attention heads, 64 hidden dimensions, LayerNorm + ELU and a dropout regularizer (0.1). Intuitively, the output is a per node embedding $o_i^t \in \mathbb{R}^64$. This structure encodes the network snapshot into a representation suitable for the RSSM.
\subsection{Per Node Recurrent Dynamics}
Canonical RSSM maintains one hidden state for the entire environment. For an N-node network, this forces N nodes observable attributes to be compressed into a single k-dimensional vector, an information bottleneck that grows worse with N. G-RSSM instead maintains N parallel states for scalability each tracking one node's dynamics with cross node attention enabling coordination with neighbors. A per node GRU update 
\begin{equation}
    \tilde{h_i^t} = \mathrm{GRU}(\mathrm{Linear}([z_i^{t-1};a_i^{t-1}])), \mspace{10mu} h_i \in \mathbb{R}^{32}
\end{equation}
receives the input of the previous stochastic state $z_i^{t-1}$ and the previous action $a_i^{t-1} \in \{0,1\}$. The weights are shared across all $N$ nodes enabling variable $N$ scalability. Cross node communication
\begin{equation}
    h_i^{t} =\mathrm{LayerNorm}(\tilde{h_i^{t}} + \mathrm{MHA}(\mathbf{\tilde{H^t}}, \mathbf{\tilde{H^t}},\mathbf{\tilde{H^t}})_i)
\end{equation}

contains $\mathbf{\tilde{H^t}} \in \mathbb{R}^{N \times 32}$ which is the all nodes' GRU outputs stacked. A four-head self attention enables capturing the inter node effects. Each node's stochastic state $z_i$ is a 64-dimensional categorical latent, factored as eight independent one-hot vectors of dimension eight following the architecture proposed in DreamerV3 \cite{hafner2024dreamer}. Prior and posterior from the canonical RSSM are now transformed into:
 \begin{equation}                                                                                                                              
    p(\mathbf{z}_i^t \mid \mathbf{h}_i^t) = \prod_{k=1}^{8} \mathrm{Cat}\!\left(\boldsymbol{\pi}_k^{\mathrm{pri}}(\mathbf{h}_i^t)\right)       \text{ and}
  \end{equation}                                                                                                                                
  \begin{equation}                                                                                                                              
    q(\mathbf{z}_i^t \mid \mathbf{h}_i^t, \mathbf{o}_i^t) = \prod_{k=1}^{8}                                                                     
  \mathrm{Cat}\!\left(\boldsymbol{\pi}_k^{\mathrm{post}}([\mathbf{h}_i^t;\, \mathbf{o}_i^t])\right)                                             
  \end{equation}
where $\pi_k \in \Delta ^ 8$ is the softmax probability simplex for the k-th categorical factor. The product $\prod$ states the 8 factors are independent given $h$ or $o$. A global state is preserved for reward/continue prediction for the state:
\begin{equation}
    s^t = \frac{1}{N_{\mathrm{alive}}}\sum_{i \in alive}\mathrm{ELU}(\mathbf{W_{s}}[\mathbf{h}_i^t; \mathbf{z}_{i}^t]) \in \mathbb{R}^{256}.
\end{equation}

\subsection{Training the World Model}

The G-RSSM is trained end to end by maximizing a variational lower bound on the log-likelihood of observed network states. Particularly, we collect state transitions (as trajectories) directly from the network simulator and train the world model. The loss decomposes into reconstruction and regularization terms:
\begin{equation}
    \mathcal{L}_{\text{WM}} = \underbrace{\mathcal{L}_{\text{pos}} + \mathcal{L}_{\text{energy}} + 0.1\mathcal{L}_{\text{adj}} + \mathcal{L}_{\text{reward}} + \mathcal{L}_{\text{cont}}}_{\mathcal{L}_{\text{recon}}} + \mathcal{L}_{\text{reg}}
\end{equation}
where $\mathcal{L_{\text{reg}}} = \beta \cdot \max(\text{KL}[q \| p],\; \eta)$. Computation and exact losses are given in Table~\ref{tab:decoders}. Position, energy and reward targets are transformed via $\mathrm{symlog}(x) = \mathrm{sign}(x)\mathrm{ln}(|x| + 1)$ and reconstructed with MSE in that space for a stable learning across quantities differ by orders of magnitude. KL is computed per node over the categorical posteriors, summed and clamped below a threshold $\eta =0.1$ to prevent posterior collapse while allowing meaningful stochastic structure. We collect 180 offline episodes across six mobility scenarios using mixed behavioral policies (random, WCA, LEACH \cite{heinzelman}) and train for 100 epochs with Adam (lr= $3 \times 10^{-4}$, batch size 32, sequence length 50, gradient clipping at 100). Model converges with total loss decreasing by 80.6\% and the continue predictor achieving 97.4\% loss reduction indicating that the model learns to predict node death timing with high fidelity. Importantly, the decoder heads are used only during training to ground the latent space. At inference, only the encoder and dynamics core are needed.

\subsection{Imagination-Based Policy Training}
Given a trained world model, we now train a policy for cluster head selection entirely through imagination. In other words, the RL training only occurs within our world model without any further network interaction. Node conditioned actor operates on per node RSSM features directly:
\begin{equation}
    \mathbf{e}_i = \mathrm{ELU}\!\bigl(\mathrm{LN}(\mathbf{W}_{\mathrm{enc}}[\mathbf{h}_i^t;\,\mathbf{z}_i^t;\,a_i^{t-1}])\bigr) \in \mathbb{R}^{64},
\end{equation}
followed by a multi head self attention layer that enables inter node coordination $\hat{\mathbf{E}} = \mathrm{LN}\!\bigl(\mathbf{E} + \mathrm{MHA}(\mathbf{E},\mathbf{E},\mathbf{E})\bigr)$ and an initial per node decision $l_i^{(0)} = \mathrm{MLP(\hat{e}_i)}$.
 \begin{figure}[t]                                                                   
  \centering                                                                          
  \includegraphics[width=0.98\columnwidth]{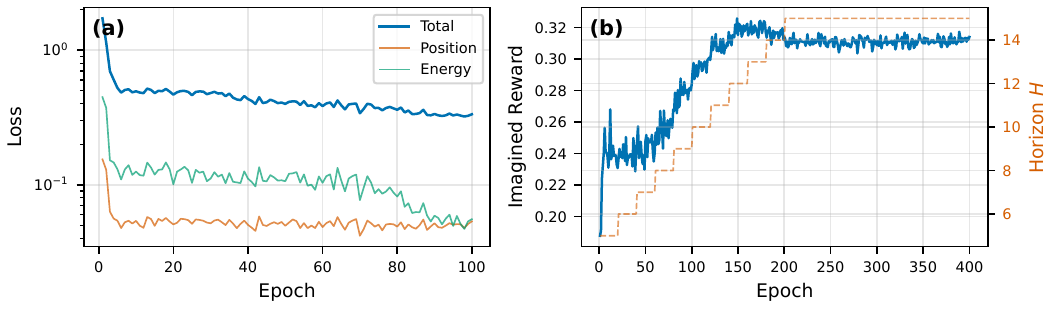}                      
  \caption{Training convergence. (a)~World model reconstruction losses over 100       
  epochs. (b)~Policy imagined reward over 400 epochs with adaptive horizon $H$.}      
  \label{fig:training}                                                                
  \end{figure}    
To enable learned coverage coordination, we introduce $R=3$ iterative refinement rounds with shared weights. At each round $k$, nodes observe their neighbors' current soft CH probabilities and adjust via $\mathbf{r}_i^{(k)} = \mathrm{ELU}\!\bigl(\mathrm{LN}(\mathbf{W}_r[\hat{\mathbf{e}}_i;\ \sigma(\ell_i^{(k)})])\bigr),$ followed by an attention $\quad \hat{\mathbf{R}}^{(k)} = \mathrm{LN}\!\bigl(\mathbf{R}^{(k)} + \mathrm{MHA}(\mathbf{R}^{(k)},\mathbf{R}^{(k)},\mathbf{R}^{(k)})\bigr)$ and finally $\ell_i^{(k+1)} = \ell_i^{(k)} + \mathrm{MLP}_r(\hat{\mathbf{r}}_i^{(k)})$ where the final CH probability becomes $p(c_i^{t} =1) = \sigma(\ell_i^{(R)})$. Each refinement round is a differentiable message passing step structurally compatible with distributed execution via local neighborhood exchanges. Imagination rollouts start from real posterior states and then proceed for $H$ steps using prior dynamics alone. In this dream state, at each imagined step the actor selects actions, reward predictor scores and continue predictor $\hat{\gamma_t}$ determines whether the rollout should continue or not. This helps explicitly learning the trajectories that cause premature node death since they are expected to generate truncated future returns. The horizon grows adaptively from $H=5$ to $H=15$ throughout the training. The actor is optimized with PPO using clipped surrogate objectives in order to maximize the objective

\begin{equation}
      J_{\text{actor}} = \mathbb{E}\bigl[\min\bigl(r_t(\theta)\hat{A}_t,\text{clip}(r_t(\theta), 1\pm\epsilon)\hat{A}_t\bigr)\bigr] + \lambda_H   
  \mathcal{H}[\pi]         \end{equation}
where $r_{t}(\theta) - \pi_{\theta}(a_t|s_t)/{\pi_{\theta_{old}}(a_t|s_t)}$, $\epsilon =0.2$ and entropy coefficient $\lambda_{H}$ is annealed from 0.03 to 0.01. Advantages are estimated via $\lambda$-returns with DreamerV3 style continue discounting:
\begin{equation}
     G_t^{\lambda} = r_t  + \hat{\gamma_{t}}((1-\lambda)V(s_{t+1} + \lambda G_{t+1}^{\lambda}), \mspace{10mu} 
\end{equation}
with $\gamma = 0.99$, $\lambda = 0.95$ where $\hat{\gamma_t}$ is the predicted continuation probability from the world model. A temporal consistency penalty $-0.1||a_t -a_{t-1}||_{1}/N$ regularizes against unnecessary CH changes between consecutive steps. The critic $V(s_t)$ is a two-layer MLP trained on the $\lambda$-return targets.

\section{Experimental Evaluation}
We conduct our experiments in a confined robust simulator aligned with the physical properties defined as main axes of a wireless ad hoc network for the clustering problem execution. Simulator and corresponding baselines are all implemented in Python, parallel to our world model implementation.

\subsection{Setup}
We employ a physics-based simulator based on simple path loss $P_{r,ij} = k_0 P_t (d_0/d_{ij})^\eta$, $\eta=3$ with Heinzelman style energy model $E_{tx}(d) = (E_{elec} + \varepsilon_{amp}d^2).\Delta t$, with an idle drain and a proportional CH overhead based on the cluster size. We evaluate 27 scenarios spanning MANET, VANET, FANET, WSN, tactical and disaster settings, with $N=30$ to 1000 nodes. We compare against six representative algorithms spanning five design paradigms: (i) ID-based (Lowest-ID \cite{lowest_id}), (ii) weight based (WCA \cite{wca}), (iii) energy aware rotation (LEACH \cite{heinzelman}), (iv) distributed iterative (HEED \cite{heed}), (v) mobility adaptive (DMAC \cite{dmac}), (vi) learning-based (DRL, a standard DQN applied to clustering). Unlike all baselines, WM-Cluster requires no online interaction after the initial data collection phase and operates on arbitrary network sizes without retraining. We evaluate four complementary aspects: CH changes, connectivity ratio defined as the fraction of alive nodes reachable by a CH, network lifetime defined as the total timesteps until connectivity drops below 50\% and Jain's fairness index of residual energies. Each configuration is evaluated over 50 episodes with identical seeds across all algorithms, statistical significance is assessed via Wilcoxon signed-rank tests with 95\% confidence intervals.
 \begin{table}[b]                                                                
    \centering                                                               
    \caption{Default scenario (50 nodes).                                                                       
    Best in \textbf{bold}, second best \underline{underlined}.}                                                                               
    \label{tab:results}                                                                                                                       
    \small                                                                                                                                    
    \setlength{\tabcolsep}{3pt}                                                                                                               
  \begin{tabular}{@{}lccccc@{}}
    \toprule                                                                                                                                  
    Algorithm & CH Chg $\downarrow$ & Conn $\uparrow$ & Life $\uparrow$ & CL-Life $\uparrow$ & Jain's $\uparrow$ \\                           
    \midrule                                                                                                       
    Lowest-ID   & \underline{507}  & \textbf{0.998} & 473 & 37.3 & 0.744 \\                                                                   
    WCA         & 1220 & \textbf{0.998} & 471 & 16.7 & 0.685 \\            
    DMAC        & 1041 & \textbf{0.998} & 453 & 18.1 & 0.725 \\                                                                               
    HEED        & 5214 & \textbf{0.998} & 429 & 3.5  & 0.709 \\                                                                               
    LEACH       & 6502 & 0.351 & 17  & 1.4  & \underline{0.925} \\                                                                            
    DRL-Cluster & 669  & 0.320 & 286 & 7.2  & 0.759 \\                                                                                        
    \midrule                                                                                                                                  
    \textbf{WM-Cluster} & \textbf{128} & 0.820 & \textbf{501} & \textbf{175.0} & \textbf{0.847} \\                                            
    \bottomrule                                                                                                                               
  \end{tabular}                                                                                                                               
                                                                                  
  \end{table}

\subsection{Does the World Model Learn Network Dynamics?}
We train the world model for 100 epochs on 180 offline episodes collected from six mobility scenarios using mixed behavioral policies. Fig.~\ref{fig:training} (a) shows the reconstruction loss decreasing from 1.72 to 0.33 (80.6\% reduction). Position and energy heads converge within 20 epochs as smooth continuous quantities while the adjacency head stabilizes more slowly due to the inherent difficulty of discrete structure prediction. The continue predictor achieves 97.4\% loss reduction (0.54 to 0.014), indicating the model accurately predicts node death timing, critical for shaping imagination rollouts.
\begin{figure*}[!t]                                                                                                                          
    \centering                                                                                                                            
    \includegraphics[width=0.75\textwidth]{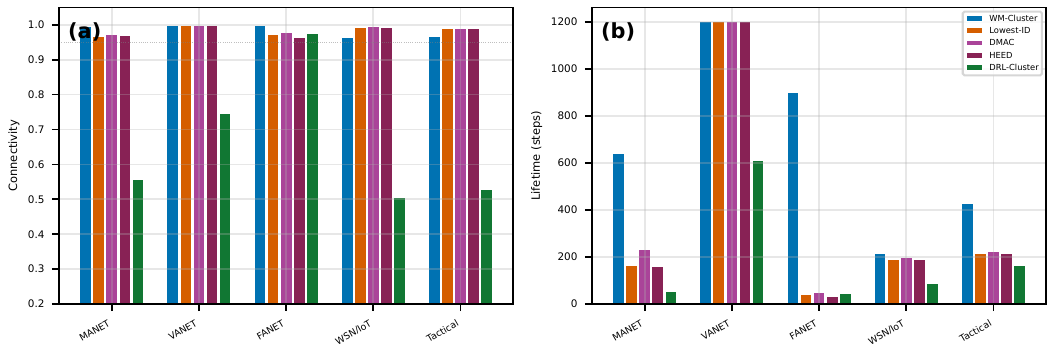}                                                                      
    \caption{Cross-scenario evaluation averaged by category (26 scenarios,                                                                    
    $N{=}30$--$1000$). (a)~Connectivity. (b)~Network lifetime.                                                                                
    WM-Cluster maintains competitive connectivity while achieving                                                                             
    substantially longer lifetime across all scenario types.}                                                                                 
    \label{fig:cross_scenario}                                                                                                                
  \end{figure*}       
\subsection{Downstream Task: Clustering Performance}
\subsubsection{Default Scenario}
Table~\ref{tab:results} summarizes performance on the default scenario (50 nodes, random waypoint, 501 steps, 50 episodes). WM-Cluster achieves 128 CH changes, 75\% fewer than Lowest-ID (507) and 81\% fewer than DRL-Cluster (669), demonstrating that imagination based training learns temporally stable cluster assignments. Cluster lifetime of 175 steps is 4.7 times longer than Lowest-ID (37.3) and 25 times longer than DRL-Cluster (7.2) enabling persistent routing structures. Energy fairness (Jain's index 0.847) exceeds all topology-based methods (Lowest-ID, DMAC) indicating the learned policy implicitly rotates the CH burden. The connectivity of 0.820 reflects a deliberate trade-off: WM-Cluster avoids over electing CHs in favor of network longevity. Topology-based methods achieve near-perfect connectivity (0.998) but exhaust the network earlier. 
\subsubsection{Cross Scenario Generalization}
WM-Cluster is trained on a single 50-node default scenario and evaluated zero-shot across 26 scenarios spanning five network categories (Fig.~\ref{fig:cross_scenario}). Despite no retraining, the learned policy achieves 0.969 average connectivity across all scenarios. It is observed that the WM-Cluster has a trend of dominance in average lifetime mainly facilitated by the reward profile it maintains.

\section{Discussion and Conclusion}
G-RSSM acts as a learned digital twin, the world model learns a compact, differentiable representation of the network that predicts major processes within ad hoc networks being a lightweight alternative to simulation-based digital twins. After collecting a certain number of trajectories from a simulator, the entire training runs without any further network interaction which may be critical for scenarios where online experimentation is infeasible. We present a learned coordination as actor refinement rounds showing that some sense of coordination can be established without heuristic coverage guarantees. Although we present G-RSSM with clustering, it is merely a proof of concept downstream application, G-RSSM can be extended to different problem formulations. Thanks to the graph based architecture, our method has the variable-$N$ property as generalization to varying network sizes. Finally, a fully distributed execution can be achieved by limiting the attention to k-hop neighbors within refinement rounds, this is left as future work.

\bibliography{references}

@article{ha2018WM,
  author       = {David Ha and
                  J{\"{u}}rgen Schmidhuber},
  title        = {World Models},
  journal      = {CoRR},
  volume       = {abs/1803.10122},
  year         = {2018},
  url          = {http://arxiv.org/abs/1803.10122},
  eprinttype   = {arXiv},
  eprint       = {1803.10122},
  bibsource    = {dblp computer science bibliography, https://dblp.org}
}

@misc{hafner2024dreamer,
      title={Mastering Diverse Domains through World Models}, 
      author={Danijar Hafner and Jurgis Pasukonis and Jimmy Ba and Timothy Lillicrap},
      year={2024},
      eprint={2301.04104},
      archivePrefix={arXiv},
      primaryClass={cs.AI},
      url={https://arxiv.org/abs/2301.04104}, 
}

@misc{wang2025dualWM,
      title={Dual-Mind World Models: A General Framework for Learning in Dynamic Wireless Networks}, 
      author={Lingyi Wang and Rashed Shelim and Walid Saad and Naren Ramakrishnan},
      year={2025},
      eprint={2510.24546},
      archivePrefix={arXiv},
      primaryClass={cs.IT},
      url={https://arxiv.org/abs/2510.24546}, 
}

@misc{zhao2025worldmodelscognitiveagents,
      title={World Models for Cognitive Agents: Transforming Edge Intelligence in Future Networks}, 
      author={Changyuan Zhao and Ruichen Zhang and Jiacheng Wang and Gaosheng Zhao and Dusit Niyato and Geng Sun and Shiwen Mao and Dong In Kim},
      year={2025},
      eprint={2506.00417},
      archivePrefix={arXiv},
      primaryClass={cs.AI},
      url={https://arxiv.org/abs/2506.00417}, 
}

@misc{wang2025worldmodelbasedlearninglongterm,
      title={World Model-Based Learning for Long-Term Age of Information Minimization in Vehicular Networks}, 
      author={Lingyi Wang and Rashed Shelim and Walid Saad and Naren Ramakrishnan},
      year={2025},
      eprint={2505.01712},
      archivePrefix={arXiv},
      primaryClass={cs.AI},
      url={https://arxiv.org/abs/2505.01712}, 
}

@inproceedings{vgrnn2019,
 author = {Hajiramezanali, Ehsan and Hasanzadeh, Arman and Narayanan, Krishna and Duffield, Nick and Zhou, Mingyuan and Qian, Xiaoning},
 booktitle = {Advances in Neural Information Processing Systems},
 editor = {H. Wallach and H. Larochelle and A. Beygelzimer and F. d\textquotesingle Alch\'{e}-Buc and E. Fox and R. Garnett},
 pages = {},
 publisher = {Curran Associates, Inc.},
 title = {Variational Graph Recurrent Neural Networks},
 volume = {32},
 year = {2019}
}

@inproceedings{
berkes2026graph,
title={Graph Dreamer: Temporal Graph World Models for Sample-Efficient and Generalisable Reinforcement Learning},
author={Ana{\"\i}s Berkes and Donna Vakalis and Yoshua Bengio and David Rolnick},
booktitle={Women in Machine Learning Workshop @ NeurIPS 2025},
year={2026},
url={https://openreview.net/forum?id=pHmgNUZixd}
}

@misc{moerland2022modelbasedreinforcementlearningsurvey,
      title={Model-based Reinforcement Learning: A Survey}, 
      author={Thomas M. Moerland and Joost Broekens and Aske Plaat and Catholijn M. Jonker},
      year={2022},
      eprint={2006.16712},
      archivePrefix={arXiv},
      primaryClass={cs.LG},
      url={https://arxiv.org/abs/2006.16712}, 
}

@misc{hafner2019learninglatentdynamicsplanning,
      title={Learning Latent Dynamics for Planning from Pixels}, 
      author={Danijar Hafner and Timothy Lillicrap and Ian Fischer and Ruben Villegas and David Ha and Honglak Lee and James Davidson},
      year={2019},
      eprint={1811.04551},
      archivePrefix={arXiv},
      primaryClass={cs.LG},
      url={https://arxiv.org/abs/1811.04551}, 
}

@misc{hafner2020dreamcontrollearningbehaviors,
      title={Dream to Control: Learning Behaviors by Latent Imagination}, 
      author={Danijar Hafner and Timothy Lillicrap and Jimmy Ba and Mohammad Norouzi},
      year={2020},
      eprint={1912.01603},
      archivePrefix={arXiv},
      primaryClass={cs.LG},
      url={https://arxiv.org/abs/1912.01603}, 
}

@misc{yang2020relationalstatespacemodelstochastic,
      title={Relational State-Space Model for Stochastic Multi-Object Systems}, 
      author={Fan Yang and Ling Chen and Fan Zhou and Yusong Gao and Wei Cao},
      year={2020},
      eprint={2001.04050},
      archivePrefix={arXiv},
      primaryClass={cs.LG},
      url={https://arxiv.org/abs/2001.04050}, 
}

@misc{hajiramezanali2020variationalgraphrecurrentneural,
      title={Variational Graph Recurrent Neural Networks}, 
      author={Ehsan Hajiramezanali and Arman Hasanzadeh and Nick Duffield and Krishna R Narayanan and Mingyuan Zhou and Xiaoning Qian},
      year={2020},
      eprint={1908.09710},
      archivePrefix={arXiv},
      primaryClass={cs.LG},
      url={https://arxiv.org/abs/1908.09710}, 
}

@misc{brody2022attentivegraphattentionnetworks,
      title={How Attentive are Graph Attention Networks?}, 
      author={Shaked Brody and Uri Alon and Eran Yahav},
      year={2022},
      eprint={2105.14491},
      archivePrefix={arXiv},
      primaryClass={cs.LG},
      url={https://arxiv.org/abs/2105.14491}, 
}

@misc{velickovic2018graphattentionnetworks,
      title={Graph Attention Networks}, 
      author={Petar Veličković and Guillem Cucurull and Arantxa Casanova and Adriana Romero and Pietro Liò and Yoshua Bengio},
      year={2018},
      eprint={1710.10903},
      archivePrefix={arXiv},
      primaryClass={stat.ML},
      url={https://arxiv.org/abs/1710.10903}, 
}

@misc{naderializadeh2020wirelesspowercontrolcounterfactual,
      title={Wireless Power Control via Counterfactual Optimization of Graph Neural Networks}, 
      author={Navid Naderializadeh and Mark Eisen and Alejandro Ribeiro},
      year={2020},
      eprint={2002.07631},
      archivePrefix={arXiv},
      primaryClass={eess.SP},
      url={https://arxiv.org/abs/2002.07631}, 
}

@INPROCEEDINGS{heinzelman,
  author={Heinzelman, W.R. and Chandrakasan, A. and Balakrishnan, H.},
  booktitle={Proceedings of the 33rd Annual Hawaii International Conference on System Sciences}, 
  title={Energy-efficient communication protocol for wireless microsensor networks}, 
  year={2000},
  volume={},
  number={},
  pages={10 pp. vol.2-},
  doi={10.1109/HICSS.2000.926982}}

@INPROCEEDINGS{wca,
  author={Chatterjee, M. and Das, S.K. and Turgut, D.},
  booktitle={Globecom '00 - IEEE. Global Telecommunications Conference. Conference Record (Cat. No.00CH37137)}, 
  title={An on-demand weighted clustering algorithm (WCA) for ad hoc networks}, 
  year={2000},
  volume={3},
  number={},
  pages={1697-1701 vol.3},
  doi={10.1109/GLOCOM.2000.891926}}

@ARTICLE{lowest_id,
  author={Lin, C.R. and Gerla, M.},
  journal={IEEE Journal on Selected Areas in Communications}, 
  title={Adaptive clustering for mobile wireless networks}, 
  year={1997},
  volume={15},
  number={7},
  pages={1265-1275},
  doi={10.1109/49.622910}}

@inproceedings{dmac,
author = {Basagni, Stefano},
title = {Distributed Clustering for Ad Hoc Networks},
year = {1999},
isbn = {0769502318},
publisher = {IEEE Computer Society},
address = {USA},
booktitle = {Proceedings of the 1999 International Symposium on Parallel Architectures, Algorithms and Networks},
pages = {310},
series = {ISPAN '99}
}

@ARTICLE{heed,
  author={Chia-Hung Lin and Ming-Jer Tsai},
  journal={IEEE Transactions on Mobile Computing}, 
  title={A Comment on "HEED: A Hybrid, Energy-Efficient, Distributed Clustering Approach for Ad Hoc Sensor Networks'}, 
  year={2006},
  volume={5},
  number={10},
  pages={1471-1472},
  doi={10.1109/TMC.2006.141}}

\end{document}